\documentclass[acmsmall,nonacm,authorversion]{acmart}

\AtBeginDocument{  \providecommand\BibTeX{{    \normalfont B\kern-0.5em{\scshape i\kern-0.25em b}\kern-0.8em\TeX}}}

\usepackage{nicefrac}
\usepackage{wrapfig}

\begin{document}

\title{A Systematic Study of Bias Amplification}

\author{Melissa Hall}
\email{melissahall@fb.com}
\affiliation{ \institution{Meta AI}
 \streetaddress{770 Broadway}
 \city{New York}
 \state{NY}
 \country{USA}
 \postcode{10003}
}
\author{Laurens van der Maaten}
\email{lvdmaaten@fb.com}
\affiliation{ \institution{Meta AI}
 \streetaddress{770 Broadway}
 \city{New York}
 \state{NY}
 \country{USA}
 \postcode{10003}
}
\author{Laura Gustafson}
\email{lgustafson@fb.com}
\affiliation{ \institution{Meta AI}
 \streetaddress{770 Broadway}
 \city{New York}
 \state{NY}
 \country{USA}
 \postcode{10003}
}
\author{Maxwell Jones}
\email{mjones2@andrew.cmu.edu}
\affiliation{ \institution{Carnegie Mellon University}
 \city{Pittsburgh}
 \state{PA}
 \country{USA}
}
\authornote{Work performed while interning at Meta AI.}
\author{Aaron Adcock}
\email{aadock@fb.com}
\affiliation{ \institution{Meta AI}
 \streetaddress{770 Broadway}
 \city{New York}
 \state{NY}
 \country{USA}
 \postcode{10003}
}

\makeatletter
\let\@authorsaddresses\@empty
\makeatother

\renewcommand{\shortauthors}{Hall, et al.}

\newcommand{\biasamp}{\textrm{BiasAmp}_{A \rightarrow T}}
\newcommand{\calA}{\mathcal{A}}
\newcommand{\calT}{\mathcal{T}}

\begin{abstract}
Recent research suggests that predictions made by machine-learning models can amplify biases present in the training data.
When a model amplifies bias, it makes certain predictions at a higher rate for some groups than expected based on training-data statistics.
Mitigating such bias amplification requires a deep understanding of the mechanics in modern machine learning that give rise to that amplification.
We perform the first systematic, controlled study into when and how bias amplification occurs.
To enable this study, we design a simple image-classification problem in which we can tightly control (synthetic) biases.
Our study of this problem reveals that the strength of bias amplification is correlated to measures such as model accuracy, model capacity, model overconfidence, and amount of training data.
We also find that bias amplification can vary greatly during training. 
Finally, we find that bias amplification may depend on the difficulty of the classification task relative to the difficulty of recognizing group membership: bias amplification appears to occur primarily when it is easier to recognize group membership than class membership.
Our results suggest best practices for training machine-learning models that we hope will help pave the way for the development of better mitigation strategies.

\noindent Code can be found at \href{https://github.com/facebookresearch/cv_bias_amplification/tree/main}{\texttt{https://github.com/facebookresearch/cv\_bias\_amplification}}.
\end{abstract}

\maketitle

\section{Introduction}
\label{sec:introduction}
Several recent studies have presented results suggesting that the predictions of machine-learning models do not just reproduce biases present in the training data, but that they can \emph{amplify} such biases as well~\cite{foulds2020intersectional, wang2021directional, zhao2017men}.
A model that amplifies bias makes certain predictions at a higher rate for some groups than is to be expected based on statistics of the training data.
Bias amplification is concerning as it can foster the proliferation of undesired stereotypes~\cite{dinan2020queens,stock2017convnets,zhao2017men,zhao2021understanding} or lead to unjustifiable differences in model accuracy between subgroups of users~\cite{boulamwini2018gendershades,devries2019everyone}.

To be able to develop methods that mitigate bias amplification, it is imperative that we understand the mechanics that lead to the amplification.
In particular, the existence of bias amplification suggests that machine-learning models are not always doing what we expect them to do: \emph{viz.}, make predictions according to the statistics present in their training data.
When does bias amplification occur and why? 
Even though several studies have proposed measures for the severity of bias amplification~\cite{foulds2020intersectional, mehrabi2022survey, wang2021directional, zhao2017men}, this question is still largely unanswered.

In this paper, we present a systematic, controlled study of bias amplification.
We design a simple image-classification task that facilitates tight control of synthetic biases.
In line with prior work~\cite{foulds2020intersectional,wang2021directional, zhao2017men}, we find that models trained for this classification task, indeed, amplify biases present in their training data.
We use the ability to control (synthetic) biases to study six key research questions (RQs) aimed at increasing our understanding of bias amplification:
\begin{itemize}
\item \emph{RQ1}: How does bias amplification vary as the bias in the data varies?
\item \emph{RQ2}: How does bias amplification vary as a function of model capacity?
\item \emph{RQ3}: How does bias amplification vary as a function of training set size?
\item \emph{RQ4}: How does bias amplification vary as a function of model overconfidence?
\item \emph{RQ5}: How does bias amplification vary during model training?
\item \emph{RQ6}: How does bias amplification vary as a function of the relative difficulty of recognizing class membership versus recognizing group membership?
\end{itemize}
We observe that bias amplification tends to increase with bias in the training set in many of our experiments.
Similarly, we observe that bias amplification varies with model capacity: models with more parameters and/or less regularization can amplify biases, but models with too few parameters and/or too much regularization can amplify biases even more.
Bias amplification also greatly varies with training set size: models trained on very small or very large training sets appear to amplify biases less.
We also find that there is a (weak) relation between model overconfidence (that is, poor calibration~\cite{guo2017calibration}) and bias amplification, and we observe that the degree of bias amplification can vary greatly during model training.
In many of our experiments, we find that the behavior of bias amplification depends on the difficulty of the classification task relative to the difficulty of group membership recognition.
For example, if it is easier to recognize the group of an image than it is to recognize the class of that image, models heavily amplify biases in the early stages of training.
But if recognizing class membership is easier than recognizing group membership, instead, bias is actually dampened in the early stages of training. 

Altogether, the results of our study provide intuitions for when bias amplification occurs and why.
Our results also suggest some best practices that may help reduce bias mitigation in real-world machine-learning models.
For example, our results suggest that careful cross-validation of hyperparameters related to model capacity, regularization, and training duration can be used to substantially reduce bias amplification of the final model.
Collecting more training data may reduce bias amplification as well.
We hope that our study helps pave the way for the development and adoption of mitigation strategies for bias amplification in machine learning.

\section{Experiments}
\label{sec:experiments}
We perform experiments designed to understand key characteristics of bias amplification.
To do so, we design an image-classification task in which each image has both a class and a group, and in which we can introduce synthetic biases by altering the group assignment of images.
Unless otherwise noted, we use the exact same setup in all experiments.

\subsection{Experimental Setup}
\label{sec:experimental_setup}
\paragraph{Classification datasets.}
We perform image-classification experiments on three image datasets: (1) the Fashion MNIST dataset~\cite{xiao2017fashionmnist}, (2) the CIFAR-10 dataset~\cite{krizhevsky2009cifar}, and (3) the CIFAR-100 dataset~\cite{krizhevsky2009cifar}.
For all datasets, we use the standard split into training and test set.
Because our analyses are easier to perform with binary classification problems, we convert the datasets to have binary labels by randomly selecting half of the classes to be positive and the other half of the classes to be negative.
We measure the accuracy of models trained to predict \emph{class membership}.
To mitigate the effect of a particular random class assignment, we average the resulting accuracy values over 20 random assignments of the original classes to the binary classes.
As bias measures are known to suffer from the Rashomon effect~\cite{breiman2001,wang2021directional}, we report both the resulting average accuracies and the corresponding $95\%$ confidence intervals.

\paragraph{Group membership.}
We consider a classification model to be \emph{biased} if it predicts a particular (binary) \emph{class} at a disproportionate prediction rate for examples from a particular \emph{group}.
When the group is, for example, a gender group, an age group, or an ethnic group, such a bias can be harmful~\cite{dinan2020queens,hendricks2018women,radford2021clip,stock2017convnets,zhao2017men,zhao2021understanding}.
In our experiments, we create two synthetic groups in such a way that groups are not causally related to classes, but the correlation between classes and groups can be controlled.
This enables us to tightly control synthetic biases in our training sets.
We also strive to make it very easy for a model to recognize group membership.
This allows us to measure precisely to what extent class predictions were influenced by group membership, that is, to assess to what extent predictions are biased.

\begin{figure}[t]
    \centering
    \includegraphics[width=0.117\textwidth]{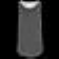}
    \includegraphics[width=0.117\textwidth]{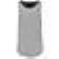}
    \includegraphics[width=0.117\textwidth]{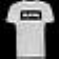}
    \includegraphics[width=0.117\textwidth]{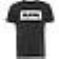}
    \hfill
    \includegraphics[width=0.117\textwidth]{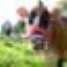}
    \includegraphics[width=0.117\textwidth]{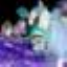}
    \includegraphics[width=0.117\textwidth]{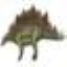}
    \includegraphics[width=0.117\textwidth]{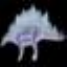}
    \caption{\textbf{Left:} Two examples of inversions performed on Fashion MNIST images. \textbf{Right:} Two examples of inversions performed on CIFAR-100 images. For each pair, the original image is on the left and the inverted image is on the right.}
    \label{fig:inversion_examples}
\end{figure}

In practice, we create two groups in our image-classification problems by \emph{inverting} some of the images in a dataset and not inverting others according to the process described below.
Using image inversion to create groups has two main advantages: (1) it hardly introduces new visual features into the images that may alter the image-classification problem and (2) it is straightforward for image-recognition models to recognize whether or not an image is inverted.\footnote{In preliminary experiments, we found that the test accuracy of a residual network trained to recognize image inversion is $100\%$ on Fashion MNIST images and $96\%$ on CIFAR-100 images.}
This allows us to tightly control the correlation between classes and groups without introducing causal relations between them.
Figure~\ref{fig:inversion_examples} shows examples of inverted and non-inverted images from the Fashion MNIST and CIFAR-100 datasets.

\paragraph{Controlling dataset bias.}
To be able to study bias (amplification) in a controlled fashion, we use the image-inversion procedure to synthetically introduce bias in our binary classification problems.
Specifically, for all images corresponding to a single task in the input dataset, we randomly select positively labeled images with rate $\nicefrac{1}{2} - \epsilon$ and invert them, and we randomly invert negatively labeled images with rate $\nicefrac{1}{2} + \epsilon$ (we choose $\epsilon \in [0, \nicefrac{1}{2}]$).
This leads to a bias of strength $2\epsilon$ in the dataset: 
If $\epsilon = 0$, image inversion (\emph{i.e.}, group membership) carries no information on whether the images has a positive or a negative label (\emph{i.e.}, class membership).
By contrast, group membership uniquely defines class membership when $\epsilon = \nicefrac{1}{2}$.
Hence, $\epsilon = 0$ corresponds to an \emph{unbiased} dataset in which group membership does not carry information about class membership, $\epsilon = \nicefrac{1}{2}$ corresponds the a \emph{fully biased} setting in which group membership uniquely determines class membership, and values of $\epsilon \in (0, \nicefrac{1}{2})$ correspond to \emph{partly biased} datasets.

\paragraph{Model training.}
Unless otherwise specified, all our models are residual networks~\cite{he2016residual} that are trained to minimize the binary cross-entropy loss between the model prediction and the true (binary) class label.
We closely follow the training procedures in~\cite{he2016residual}. 
Specifically, we train our models using mini-batch stochastic gradient descent (SGD) with a Nesterov momentum~\cite{nesterov1983} of $0.9$ for $500$ epochs.
The models are trained using weight decay ($\ell_2$-regularization) with a decay parameter of $10^{-4}$.
We warm up the training by setting the learning rate to $0.01$ for one epoch \cite{goyal2017hour}.
Subsequently, the learning rate is set to $0.1$ and decayed twice by a factor of $10$ after $250$ and $375$ epochs, respectively. 
We train our models on a single GPU using a batch size of $128$.

During training, we adopt the data augmentation procedure of \cite{he2016residual} by: (1) randomly cropping training images, (2) flipping the resulting image horizontally with probability $\nicefrac{1}{2}$, and (3) resizing the crops to size $28 \times 28$ pixels (for Fashion MNIST) or $32 \times 32$ pixels (for CIFAR-10 and CIFAR-100).
No data augmentation is used at test time.
We normalize all images before inputting them into the model by subtracting a per-channel mean value and dividing by a per-channel standard deviation; we compute the mean and standard deviation values on the original input images (without inversion or data augmentation).
When training models on CIFAR-10 and CIFAR-100, we follow \cite{he2016residual} and pad the images with zeros before inputting them into the model. 
We did not use such padding when training models on Fashion MNIST.

\paragraph{Bias amplification measure.}
To measure the severity of bias amplification in a model, we adopt the directional bias amplification measure $\biasamp$ from Wang and Russakovsky~\cite{wang2021directional}.
We selected this measure because it naturally disambiguates different types of bias amplification and because it accounts for varying base rates of group membership. 
We only give a concise treatise of the measure here, and refer the reader to~\cite{wang2021directional} for further details.

Suppose we have a set of \emph{classes}, $\calT$, and a set of \emph{groups}, $\calA$.
In our setup, $\calT = \{ -1, +1 \}$ and $\calA = \{\textrm{inverted}, \textrm{not inverted}\}$, where the binary labels $t \in \calT$ were obtained by the random class assignment described above.
The $\biasamp$ measure defines \emph{bias} as a difference in the prevalence of a class label $t \in \calT$ between groups $a \in \calA$.
For example, bias is present if inverted images are more likely to be positively labeled.
Bias can be present in the class labels of the dataset and/or in the (hard) predictions made by the classification model.
Denote by $Pr(T_t = 1)$ the probability that an example in the dataset has class label $t$, and by $Pr(\hat{T}_t = 1)$ the probability that an example in the dataset is labeled as class $t$ by the model.
With these definitions in place, \cite{wang2021directional} defines \emph{bias amplification} as the difference in bias between the labels in the dataset and the labels predicted by the model:
\begin{equation}
\biasamp = \frac{1}{| \calA | | \calT |} \sum_{a \in \calA, t \in \calT} y_{at} \Delta_{at} - (1 - y_{at}) \Delta_{at}.
\end{equation}
Herein, $\Delta_{at}$ measures the difference between the bias in the dataset and the bias in the model predictions:
\begin{equation}
\Delta_{at} = Pr(\hat{T}_t = 1| A_a = 1) - Pr(T_t = 1| A_a = 1).
\end{equation}
In the definition of $\biasamp$, $y_{at}$ alters the sign of the difference $\Delta_{at}$ to correct for the fact that the bias can have two directions.
Specifically, $y_{at} \in \{0, 1\}$ is a binary variable that indicates the direction of the bias:
\begin{equation}
y_{at} = \left[ Pr(T_t = 1, A_a = 1) > Pr(T_t = 1) Pr(A_a = 1) \right],
\end{equation}
where $[ \dots ]$ are Iverson brackets.
In all our experiments, we compute $\biasamp$ by measuring both $Pr(\hat{T}_t = 1)$ and $Pr(T_t = 1)$ on the test set after training the model on the training set.
The train and test datasets come from the same distribution.

The value of $\biasamp$ is $0$ if the model predictions are exactly as biased as the labels in the dataset.
Note that this does not imply that the model predictions are unbiased: it only implies that the model does not \emph{amplify} the bias present in the dataset.
If $\biasamp$ is negative, the model predictions dampen the bias present in the dataset.
By contrast, a positive $\biasamp$ value indicates that the model predictions amplify the bias in the dataset.

\subsection{Results}
\label{sec:results}
We use the setup described above to perform experiments that aim to answer our six research questions.
We present the results of those experiments organized by research question (RQ).

\begin{figure}[t]
    \centering
    \includegraphics[width=0.48\textwidth]{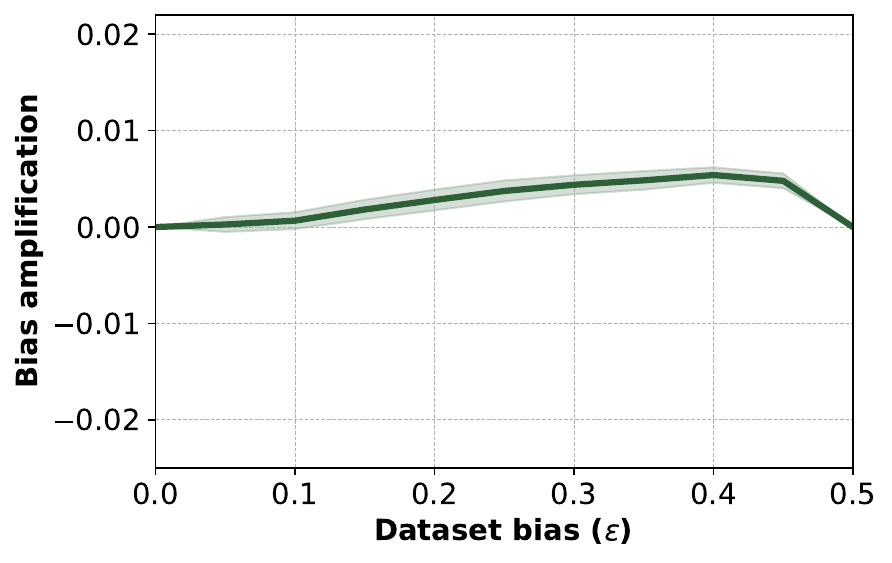}
    \includegraphics[width=0.48\textwidth]{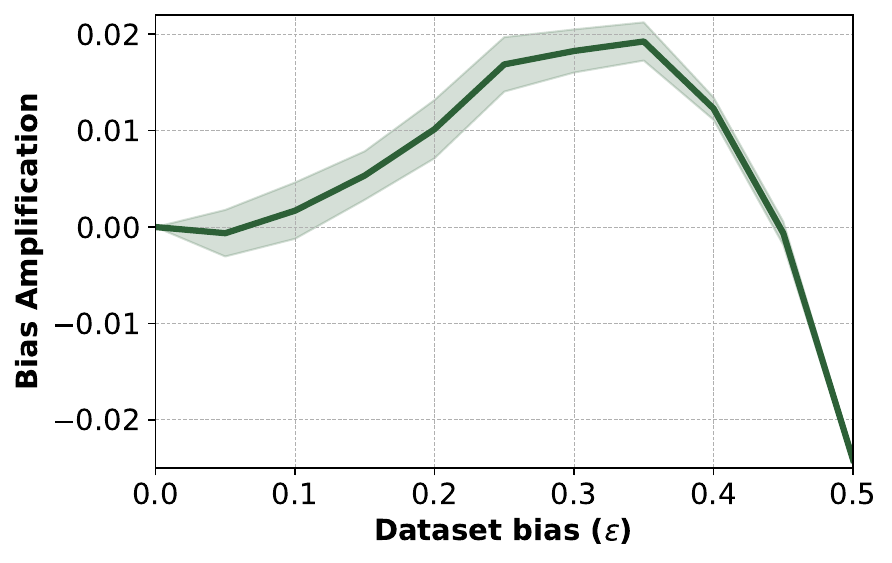}
    \caption{Bias amplification, $\biasamp$, as a function of the degree of bias, $\epsilon$, for (\textbf{left}) ResNet-18 models trained on the Fashion MNIST dataset and (\textbf{right}) ResNet-110 models trained on the CIFAR-100 dataset. Shaded regions indicate the 95\% confidence intervals over 20 independent experiments.}
    \label{fig:biasamp_epsilon}
\end{figure}

~\\
\paragraph{\bf RQ1: How does bias amplification vary as the bias in the data varies?}~\\
We perform experiments in which we vary the amount of bias in the Fashion MNIST dataset by generating training and test sets with different levels of bias, \emph{i.e.}, by varying $\epsilon$.
We train residual networks with 18 layers (ResNet-18) on these training sets, and use the corresponding test sets to measure the bias amplification, $\biasamp$, of the trained networks. 
We report the mean bias amplification values and the corresponding 95\% confidence intervals as a function of the amount of bias, $\epsilon$, in the training set.
The results of this experiment are presented in the left pane of Figure~\ref{fig:biasamp_epsilon}.

The results in the figure demonstrate a clear relationship between the amount of bias in the training set and the degree to which a trained model \emph{amplifies} that bias.
When the training set is unbiased ($\epsilon=0$), no bias amplification occurs because no bias is present.
Similarly, no bias amplification occurs when the training set is fully biased ($\epsilon=\nicefrac{1}{2}$) because it is impossible to amplify an already maximum bias.
However, for intermediate $\epsilon$ values (\emph{i.e.}, in partially biased training sets), the trained models consistently amplify the bias present in the training data.
Bias amplification generally \emph{increases} with the amount of bias in training data, until the bias in the data is nearly maximized ($\epsilon=0.5$).

We repeat the same experiment on the CIFAR-100 dataset with a 110-layer residual network (ResNet-110).
We present the results of that experiment in the right pane of Figure~\ref{fig:biasamp_epsilon}.
The results in the figure show a similar pattern: bias amplification is consistently present for intermediate $\epsilon$ values, though peak amplification happens for a lower $\epsilon$ value on CIFAR-100 than on Fashion MNIST.
A notable difference between the results on Fashion MNIST and those on CIFAR-100 is that bias amplification is negative when the CIFAR-100 dataset is maximally biased ($\epsilon=\nicefrac{1}{2}$).
We surmise that this happens because the group membership of CIFAR-100 images (\emph{i.e.}, whether or not the image was inverted) cannot always be recognized correctly by a model.\footnote{By contrast, the uniform background color or Fashion MNIST images makes it trivial for a model to recognize whether or not they were inverted.}
To obtain zero bias amplification at $\epsilon=\nicefrac{1}{2}$, a model needs to be a perfect predictor of group membership.
Hence, when the model incorrectly recognizes the group membership of some of the images, a negative bias amplification (\emph{i.e.}, bias dampening) is obtained.

~\\
\paragraph{\bf RQ2: How does bias amplification vary as a function of model capacity?}~\\
It is well-known that the capacity of machine-learning models influences their classification performance: models with higher capacity generally achieve a lower classification error on the training set, but overfitting effects may lead them to attain higher classification errors on held-out test data if insufficient training data is available, the model is not sufficiently regularized, \emph{etc.}
To understand how model capacity impacts bias amplification, we perform experiments in which we measure bias amplification while adjusting the capacity of our models.
We adjust model capacity in three ways: (1) via the \emph{depth} of the model, that is, the number of layers in the convolutional network; (2) via the \emph{width} of the model, that is, the number of channels in each convolutional network layer; and (3) via the \emph{regularization} of the model, that is, the amount of weight decay to the model during training.

\begin{figure}
  \centering
  \includegraphics[width=0.32\textwidth]{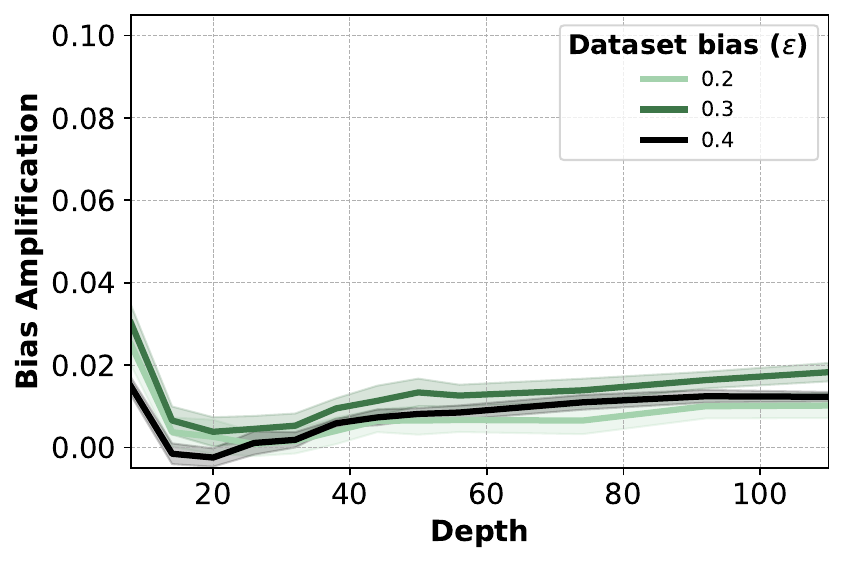}
  \includegraphics[width=0.32\textwidth]{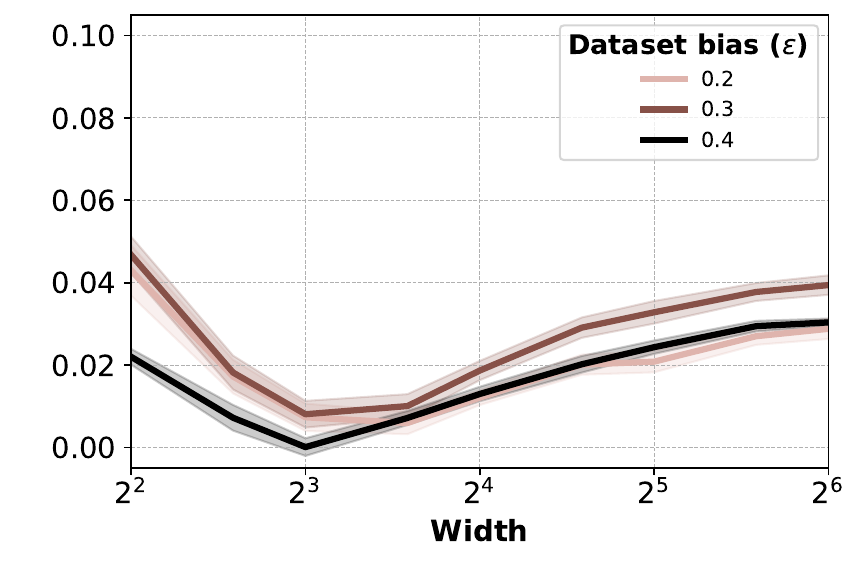}
  \includegraphics[width=0.32\textwidth]{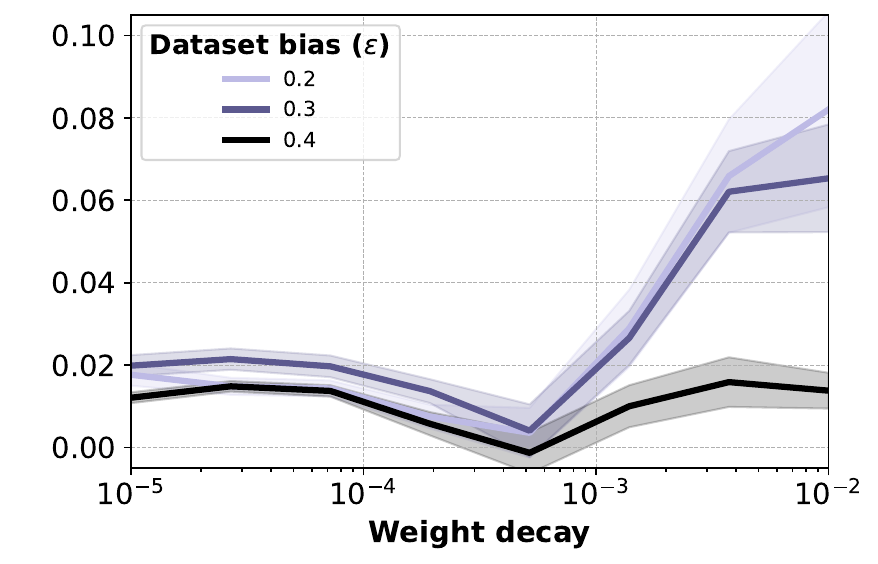}
  \includegraphics[width=0.32\textwidth]{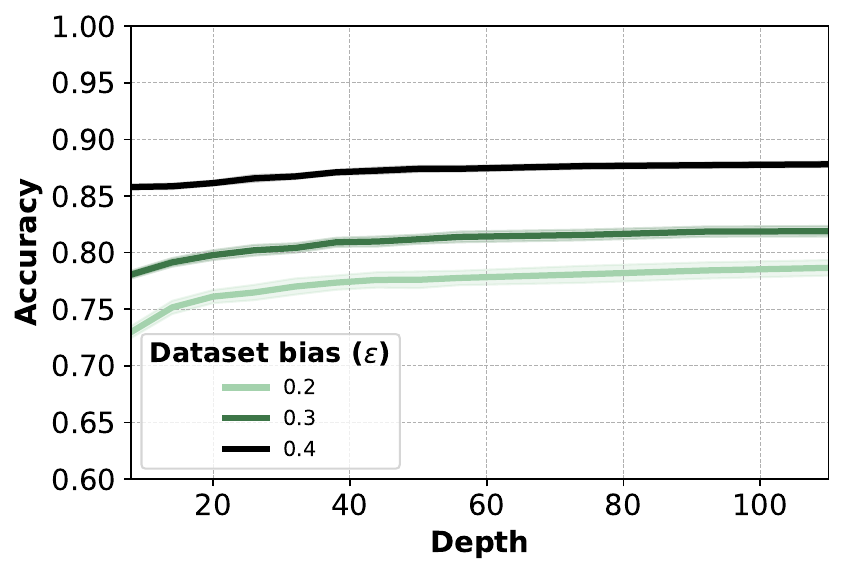}
  \includegraphics[width=0.32\textwidth]{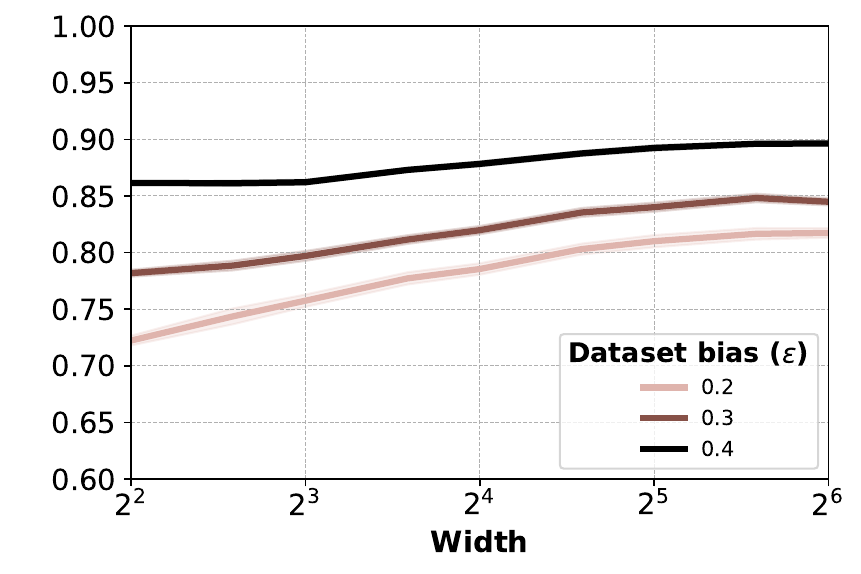}
  \includegraphics[width=0.32\textwidth]{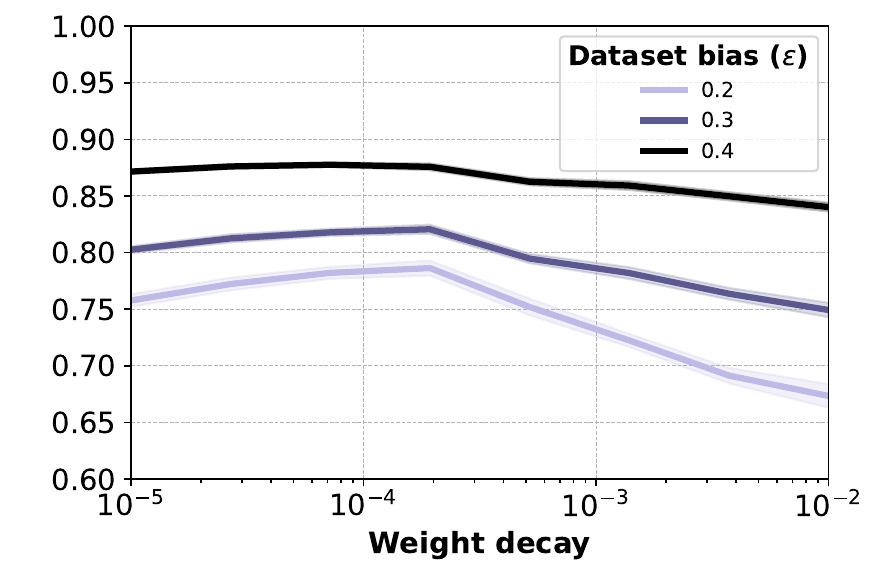}
  \caption{Bias amplification (\textbf{top}) and accuracy (\textbf{bottom}) on the CIFAR-100 dataset as a function of three measures of model capacity. Each line represents a different amount of bias ($\epsilon$) in the training set. Shaded regions indicate the $95\%$ confidence intervals across 20 models. \textbf{Left:} Results for varying model depths.  \textbf{Middle:} Results for varying model widths.  \textbf{Right:} Results for varying weight decays.}
  \label{fig:biasamp_and_capacity}
\end{figure}

We focus on the CIFAR-100 dataset in this set of experiments because CIFAR-100 images are harder to classify than Fashion MNIST images: this makes it more likely that models with different capacities will produce substantially different predictions.
We use the ResNet-110 model from our RQ1 experiments as our base model.
This model has a depth of 110 layers and a width of 16 channels in the first layer; it regularizes model parameters during training using a weight decay of $10^{-4}$.
We vary the depth, width, and weight decay of this base model in three separate experiments, leaving all other parameters fixed.
Specifically, we experiment with depths that range between $8$ and $110$, with widths ranging between $4$ and $64$, and with logarithmically spaced weight decay values between $10^{-5}$ to $10^{-2}$.
As before, we vary the dataset bias, $\epsilon$, between $0$ and $\nicefrac{1}{2}$.\footnote{To make the plots clearer, we only show values of $\epsilon$ between $0.2$ and $0.4$. As before, we found no bias amplification for $\epsilon=0$.}
The top row of Figure~\ref{fig:biasamp_and_capacity} shows the results of these experiments.

Irrespective of whether we vary model depth, width, or weight decay, the results suggest that bias amplification follows a ``v-shape'': it increases when model capacity increases beyond a certain level, but it also increases when model capacity is reduced below a certain level.
We surmise there are different explanations for these two increases.
When the capacity of a model is limited, it needs to rely on features that are easy to extract when making class predictions.
When the dataset is biased ($\epsilon > 0$), the model thus relies on image inversion, which is easy to recognize, in its class predictions.
This explains why bias amplification is relatively large when the model has low capacity.\footnote{We note that this explanation relies on the assumption that \emph{group membership} features are relatively easy to extract and, hence, that our observations may change had we not used image inversion to construct our synthetic groups. We investigate how the difficulty of recognizing group membership influences bias amplification in RQ6.}
In contrast, when the capacity of a model is large, bias amplification may increase because the model has the capacity to extract both features that indicate class membership and features that indicate group membership.
This allows the model to use group membership features to increase the confidence of its predictions, which reduces the training loss.\footnote{Indeed, we surmise the increase in bias amplification in very high-capacity models is related to the tendency of such models to be overconfident~\cite{guo2017calibration}; we investigate this relation further in RQ4.}

The relation between model capacity and bias amplification resembles the well-known relation between model capacity and \emph{generalization error}.
Models with insufficient capacity have high generalization error because they cannot model the data distribution well, whereas high-capacity models may have high generalization error due to \emph{overfitting}.
Our results suggests that there exists a model-capacity ``sweet spot'' in which bias is minimally amplified, akin to model-capacity sweet spot that minimizes generalization error (for a given training set).

To investigate whether the optimal model capacities for bias amplification and generalization error coincide, we plot the test accuracy of our models in the bottom row of Figure~\ref{fig:biasamp_and_capacity}.
Test accuracy increases monotonically with model depth and width, suggesting that the (overall) optimal model is larger than the range of models we experimented with.
However, we do observe that a weight decay of $1.9\cdot10^{-4}$ appears optimal for test accuracy.
This weight-decay value is smaller than the value that minimizes bias amplification ($5.2\cdot10^{-4}$), which suggests that model designers may sometimes have to trade off bias amplification and accuracy when tuning hyperparameters.

~\\
\paragraph{\bf RQ3: How does bias amplification vary as a function of training set size?}~\\
It is well-established that the error of machine-learning models can be reduced by increasing the amount of training data (as it reduces the estimation error~\cite{boucheron2005,vapnik1982estimation}).
This raises the obvious question if bias amplification varies with training set size as well.
To answer this question, we perform experiments in which we train ResNet-110 models on stratified subsamples of the CIFAR-100 training set.
We vary the size of the subsamples to be a proportion, $p \in [0.1, 1.0]$, of the original training set.
We increase the number of training epochs by a factor of $\nicefrac{1}{p}$ so that each model performs the same number of parameter updates during training. 
We do not alter any of the other hyperparameters.

\begin{figure}[t]
    \centering
    \includegraphics[width=0.48\textwidth]{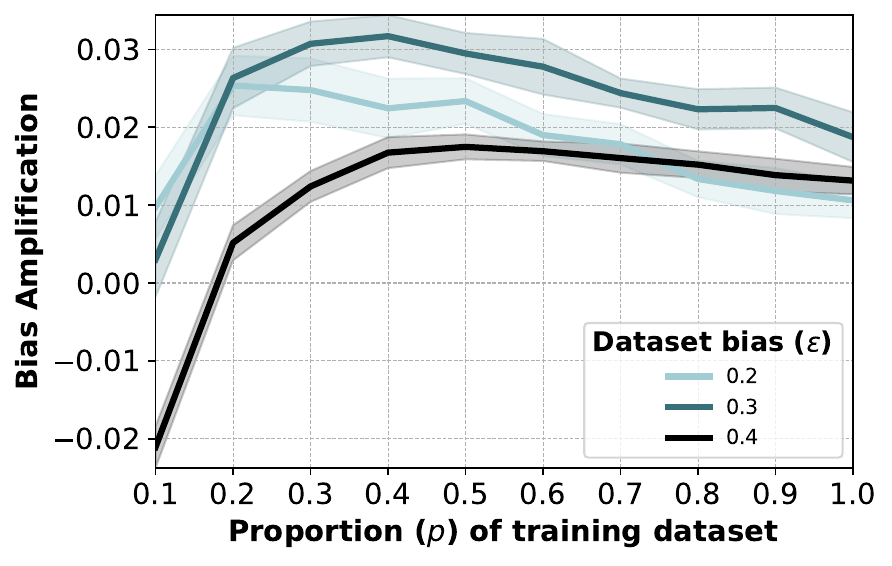}
    \includegraphics[width=0.48\textwidth]{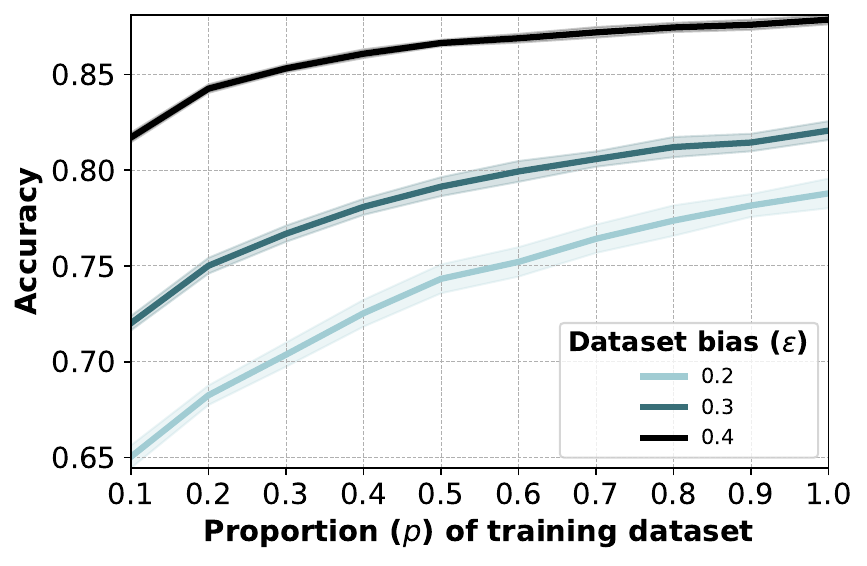}
    \caption{Bias amplification (\textbf{left}) and accuracy (\textbf{right}) of ResNet-110 models on the CIFAR-100 dataset as a function of the proportion of the training set used for training the models. The number of epochs for each model is scaled depending on the amount of training data used. Shaded regions indicate the $95\%$ confidence intervals across 20 models.}
    \label{fig:biasamp_and_training_samples}
\end{figure}

Figure~\ref{fig:biasamp_and_training_samples} shows the results of our experiments.
Whereas model accuracy increases monotonically with training set size, bias amplification varies in a more complex way.
Beyond a certain training set size, bias amplification decreases with more training data.
This is unsurprising: the additional training examples enable more accurate modeling of the data distribution, which reduces bias amplification.
Somewhat surprisingly, bias amplification is also reduced when the training set becomes very small.
We surmise this observation is due to overfitting: when trained on a small dataset, models tend to learn spurious correlations in that dataset rather than true statistical patterns such as the biases that exist in our training sets.
The model cannot amplify bias if it is unable to capture that bias in the first place.

~\\
\paragraph{\bf RQ4: How does bias amplification vary as a function of model overconfidence?}~\\
Our observation that models with higher capacity amplify bias more is reminiscent of observations that higher-capacity models tend to be more miscalibrated~\cite{guo2017calibration}.
If high-capacity models are not explicitly calibrated, they are often overconfident in the sense that the accuracy of predictions that they make with, say, $90\%$ confidence is lower than $90\%$.
We perform experiments to investigate if bias amplification is correlated to such model overconfidence.

To do so, we measure the overconfidence of our models in terms of the expected calibration error (ECE;~\cite{naeini2015}).
The ECE measures the expected value of the (absolute) difference between the model accuracy and the model confidence:
\begin{equation}
ECE(\hat{P}) =  \mathbb{E} \left[ | Pr(\hat{Y} = y | \hat{C} = c) - c | \right],
\end{equation}
where $\hat{Y}$ and $\hat{C}$ are random variables indicating the class label of an example and the model-prediction confidence for that same example, respectively, and the expectation is over all possible confidence values $c \in [0,1]$.
Because we only have access to a finite number of samples of the distribution $p(\hat{C})$, we approximate the expected value by binning $p(\hat{C})$ into 15 values and averaging those values, weighted by the number of examples per bin.
A higher ECE value indicates a larger discrepancy between the prediction confidence values and the corresponding accuracies, \emph{i.e.}, a higher degree of model overconfidence.

\begin{wrapfigure}{r}{0.5\textwidth}
    \centering
    \includegraphics[width=0.5\textwidth]{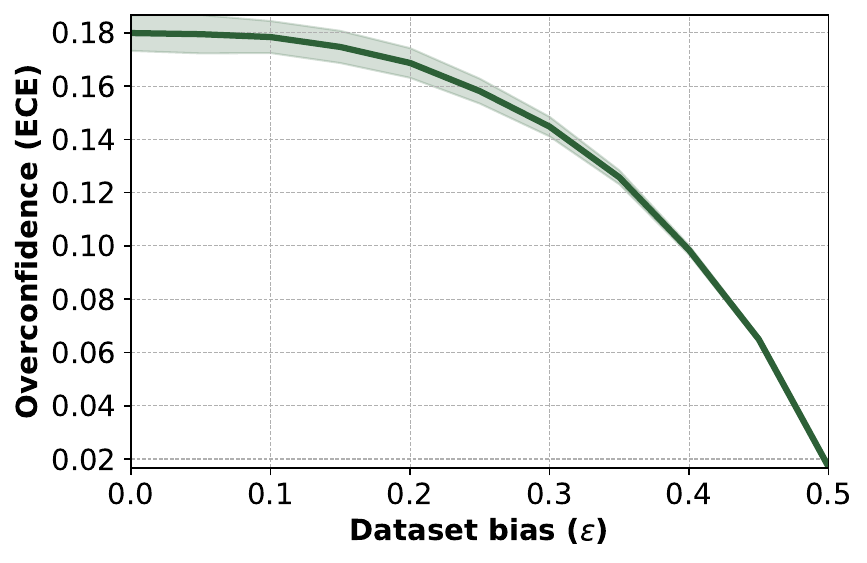}
    \caption{Expected calibration error (ECE) of ResNet-110 models on the CIFAR-100 dataset as a function of dataset bias, $\epsilon$. Shaded regions indicate the 95\% confidence intervals across 20 models.}
    \label{fig:m1_e5_ece_epsilon_cifar110}
\end{wrapfigure}

Figure~\ref{fig:m1_e5_ece_epsilon_cifar110} shows ECE as a function of the bias in the dataset, $\epsilon$, for ResNet-110 models on CIFAR-100.
We observe that model overconfidence decreases with bias in our experiment, because the task becomes easier as bias increases: if a task is very easy, a model is generally less overconfident as it correctly predicts nearly every example.

\begin{figure}[t]
\begin{minipage}{0.32\textwidth}
    \includegraphics[width=\textwidth]{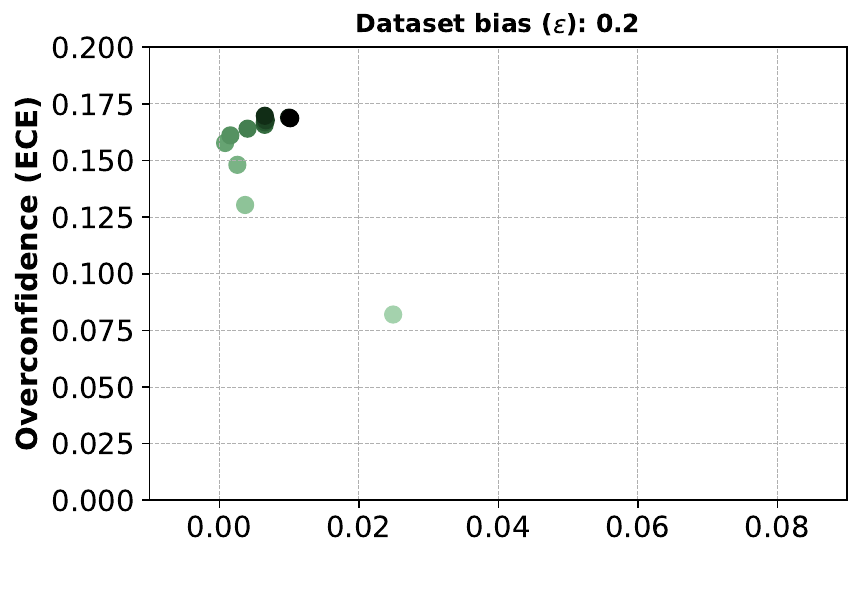}
    \includegraphics[width=\textwidth]{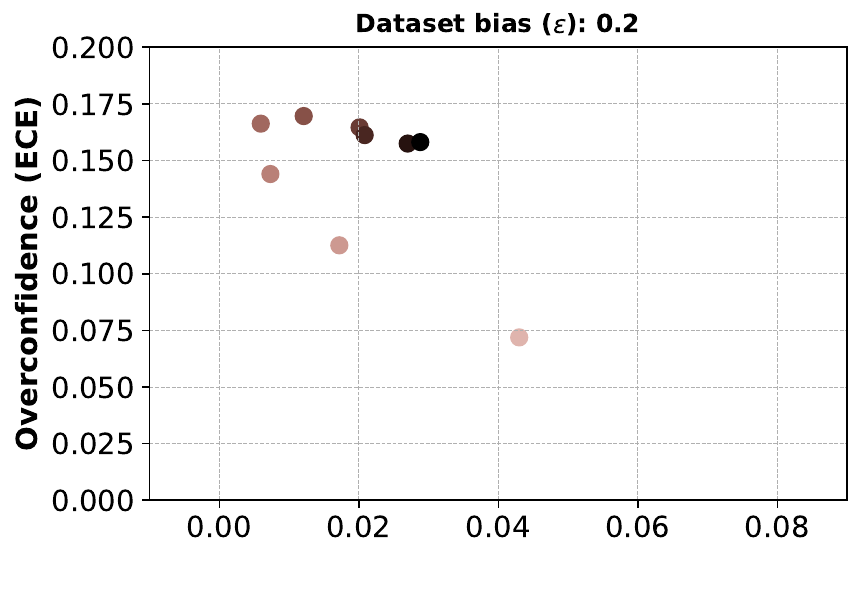}
    \includegraphics[width=\textwidth]{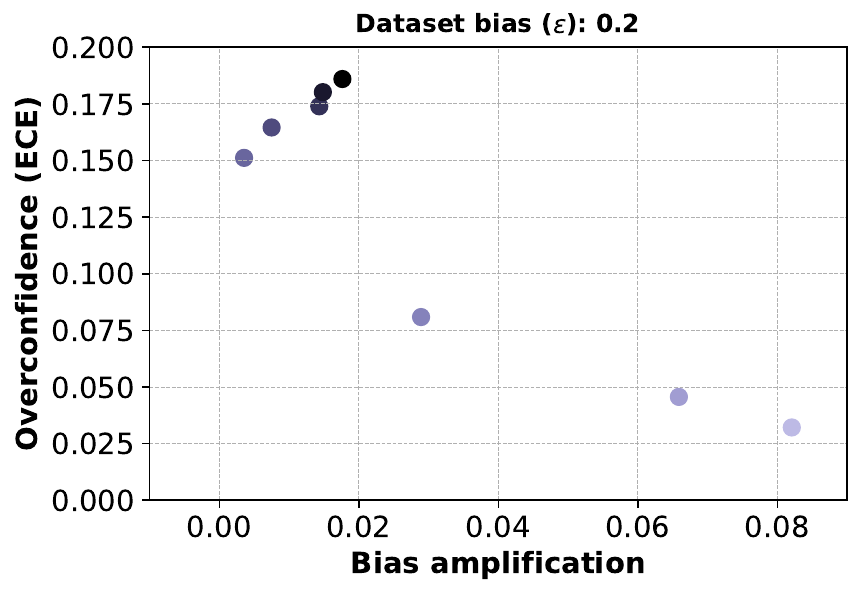}
\end{minipage}
\begin{minipage}{0.32\textwidth}
    \includegraphics[width=\textwidth]{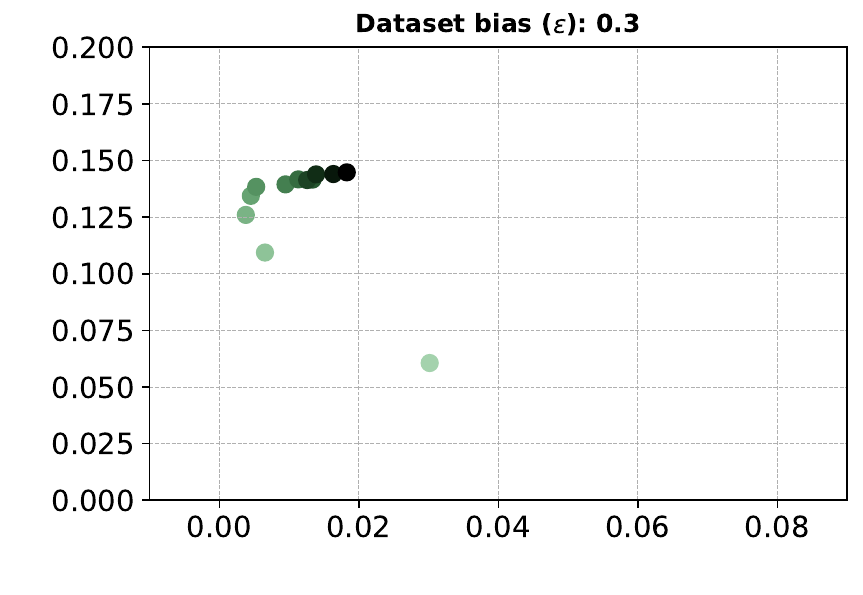}
    \includegraphics[width=\textwidth]{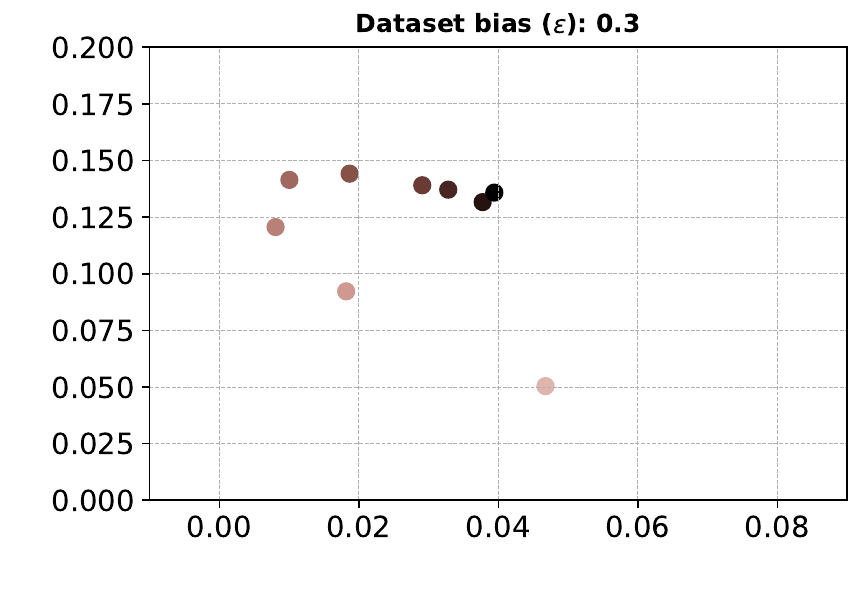}
    \includegraphics[width=\textwidth]{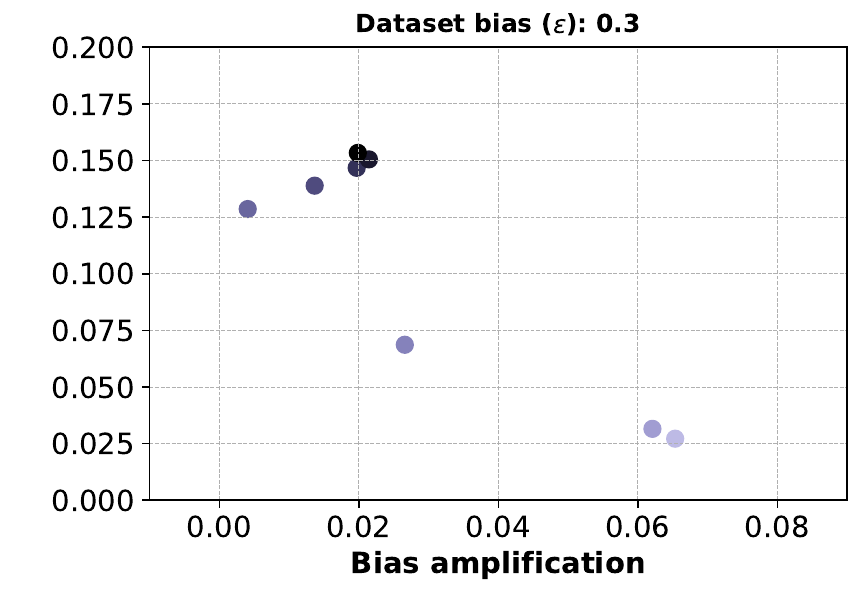}
\end{minipage}
\begin{minipage}{0.32\textwidth}
    \includegraphics[width=\textwidth]{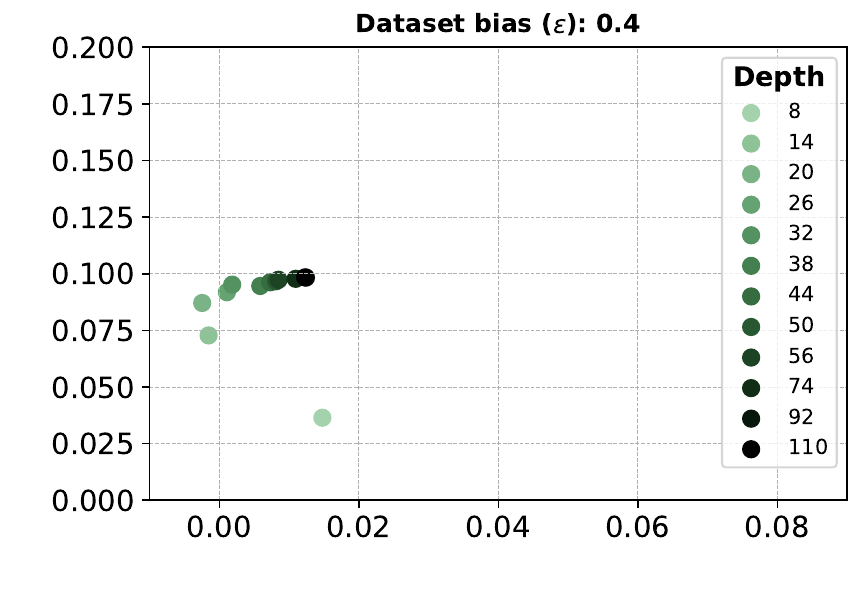}
    \includegraphics[width=\textwidth]{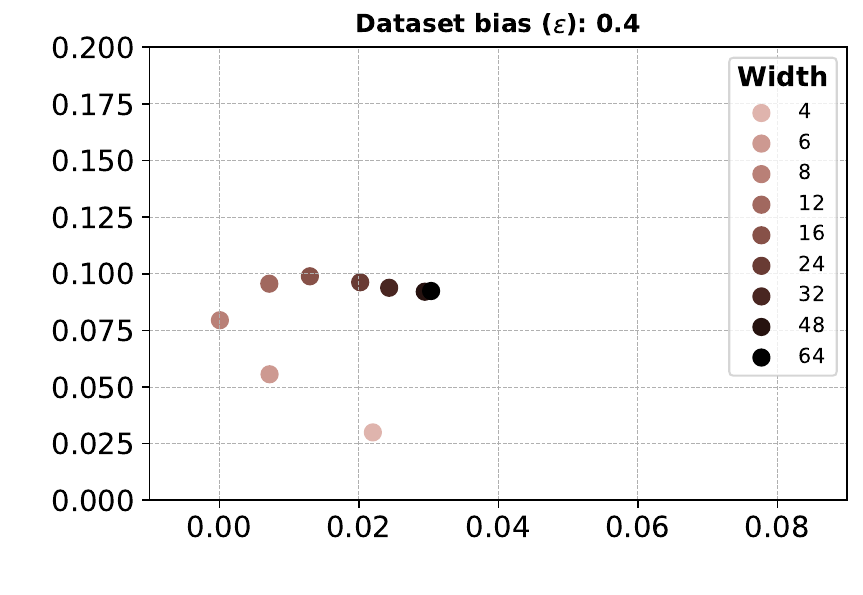}
    \includegraphics[width=\textwidth]{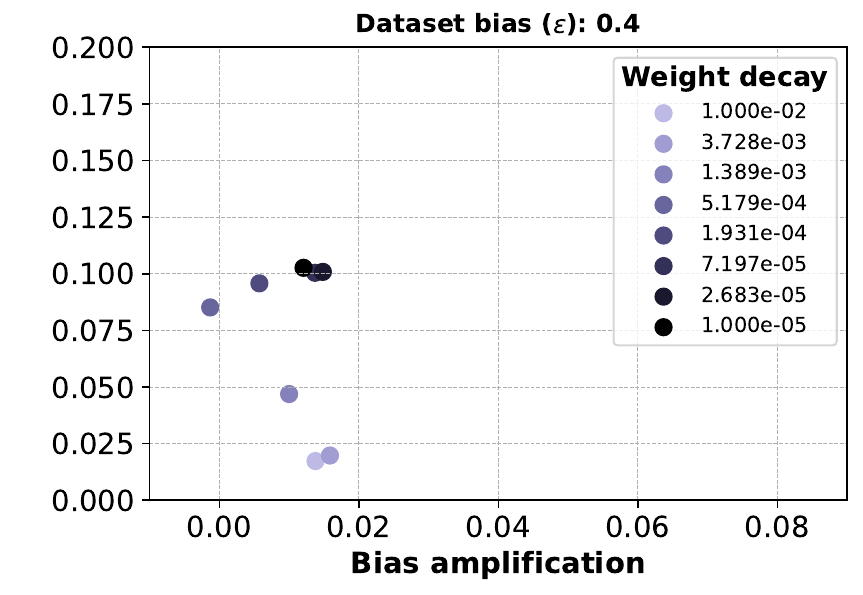}
\end{minipage}
\caption{Bias amplification and expected calibration error (ECE) of ResNet models of varying depth (\textbf{first row}), width (\textbf{second row}), and weight decay (\textbf{third row}) on the CIFAR-100 dataset, for three values of the dataset bias, $\epsilon$. Results are averaged over 20 runs.}
\label{fig:m1_e5_ece_biasamp_eps_cifardepth}
\end{figure}

Next, we study the relation between overconfidence and bias amplification by varying the capacity of the model.
Figure~\ref{fig:m1_e5_ece_biasamp_eps_cifardepth} shows this relation for three levels of dataset bias, $\epsilon$, and for three model-capacity measures: depth, width, and weight decay.
Darker points in the figure correspond to higher-capacity models.
The results show that bias amplification initially decreases as model overconfidence increases (for low-capacity models), but that bias amplification and overconfidence both increase for higher-capacity models.

\begin{figure}[t]
    \begin{minipage}{0.32\textwidth}
        \includegraphics[width=\textwidth]{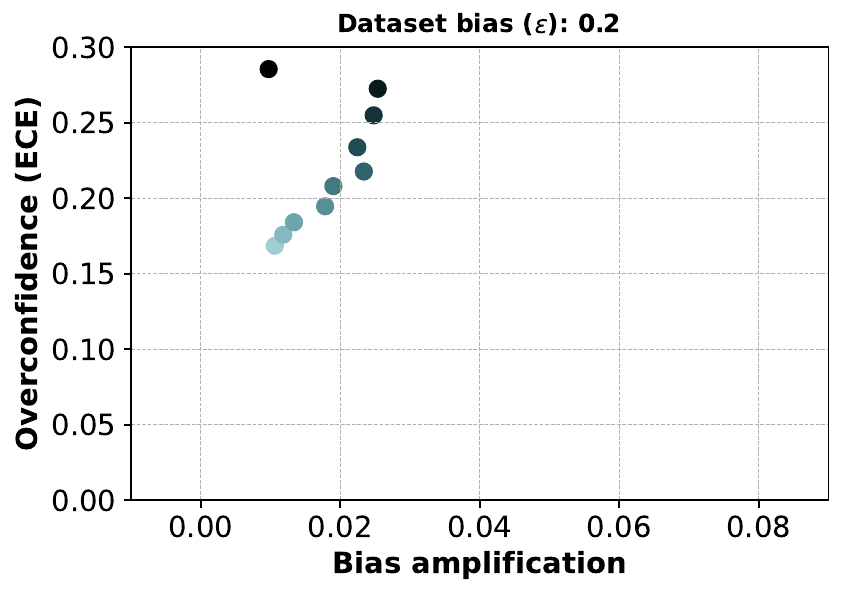}
    \end{minipage}
    \begin{minipage}{0.32\textwidth}
        \includegraphics[width=\textwidth]{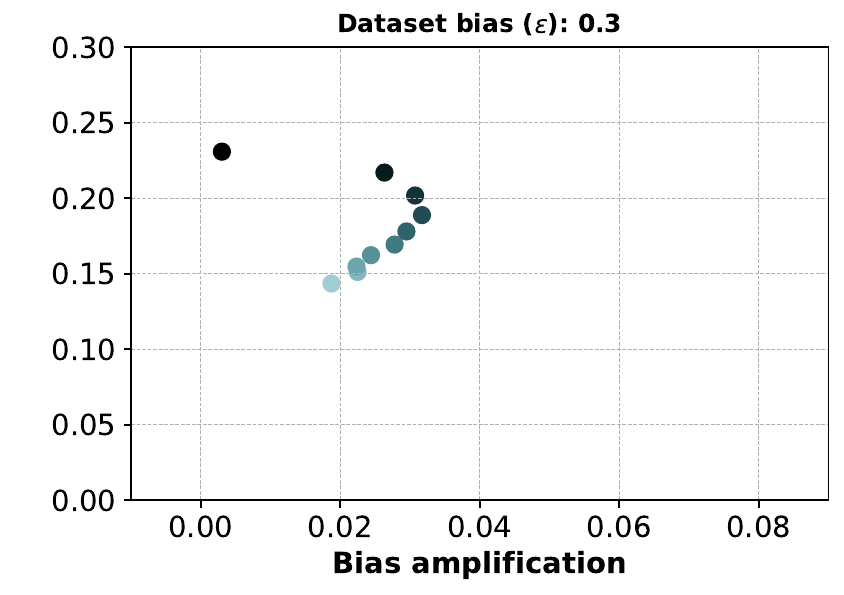}
    \end{minipage}
    \begin{minipage}{0.32\textwidth}
        \includegraphics[width=\textwidth]{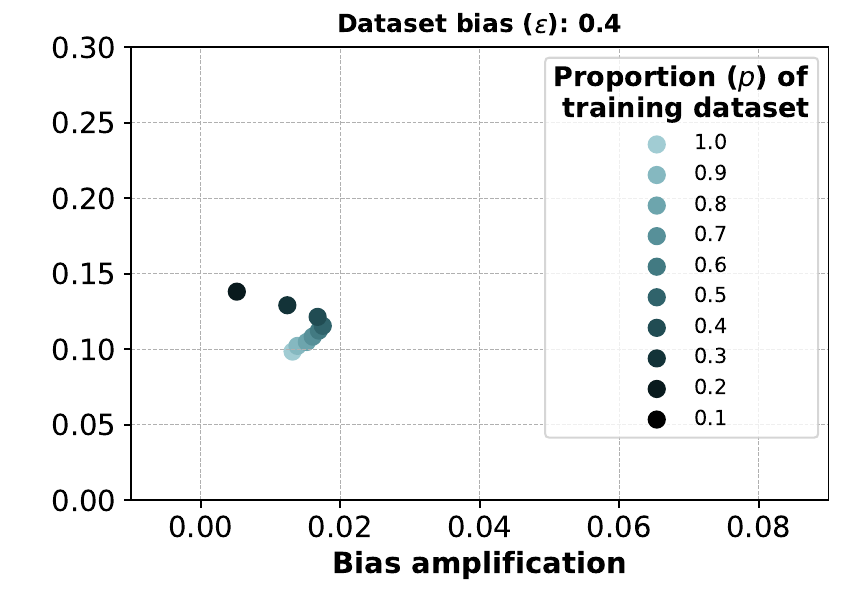}
    \end{minipage}
    \caption{Bias amplification and expected calibration error (ECE) of ResNet models of varying training dataset size on the CIFAR-100 dataset, for three values of the dataset bias, $\epsilon$. Results are averaged over 20 runs.}
    \label{fig:m1_e5_ece_biasamp_eps_cifartrainingsamples}
\end{figure}

Finally, Figure~\ref{fig:m1_e5_ece_biasamp_eps_cifartrainingsamples} studies the relationship between bias amplification and overconfidence as the size of the training set changes.
Darker points in the figure correspond to smaller training sets.
As expected, reducing the number of training examples increases the model's overconfidence (ECE). 
Bias amplification, however, initially increases as the training set size decreases but decreases again when the training set becomes very small.

\begin{figure}[t]
    \centering
    \includegraphics[width=0.51\textwidth]{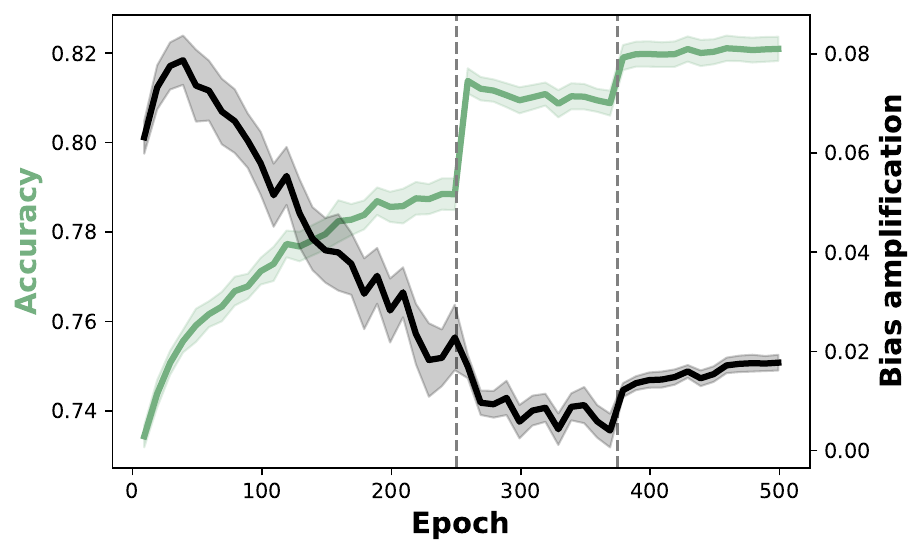}
    \hfill
    \includegraphics[width=0.46\textwidth]{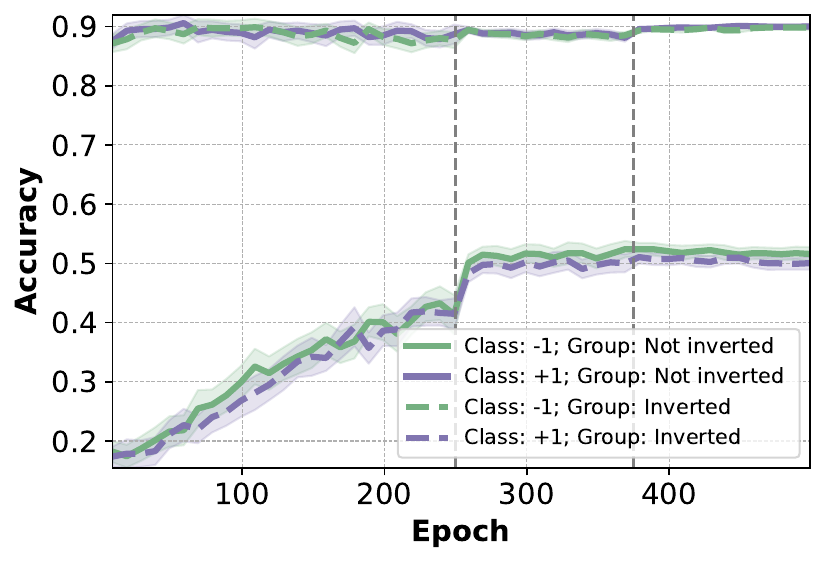}
    \caption{\textbf{Left:} Bias amplification and accuracy of ResNet-110 models during training on the CIFAR-100 dataset with a bias of $\epsilon=0.3$. \textbf{Right:} Accuracy of the same models' per class-group combination. Shaded regions indicate the 95\% confidence intervals across 50 models. Vertical dashed lines indicate epochs at which the learning rate of the mini-batch SGD optimizer is decreased.}
        \label{fig:biasamp_during_training}
\end{figure}

\paragraph{\bf RQ5: How does bias amplification vary during model training?}~\\
In our experiments so far, we have only measured bias amplification of models that were trained until convergence for $500$ epochs.
While it is feasible to train models until convergence on small datasets like Fashion MNIST or CIFAR-100, it may not be practical to do so on very large training sets.
This begs the question whether or not the degree of bias amplification of a model varies during training.
To answer this question, we measure bias amplification during the training of ResNet-110 models on a version of the CIFAR-100 dataset with bias $\epsilon=0.3$. 
The left pane of Figure~\ref{fig:biasamp_during_training} plots bias amplification as a function of training epoch in this setting; it also shows the test accuracy of the corresponding models.
The results in the figure show that bias amplification \emph{varies greatly} during training.
In particular, models amplify biases much more strongly in the early stages of training.
Bias amplification gradually declines as training proceeds and the recognition accuracy of the model increases.
However, the bias amplification increases again slightly in the final stages of training, in particular, after the learning rate is decreased to its smallest value.
Notably, bias amplification appears to increase slightly every time the learning rate is decreased.

To better understand what drives these changes in bias amplification during training, we disaggregate the model's test accuracy into the four group-task combinations in the right pane of Figure~\ref{fig:biasamp_during_training}.
We observe that the model very quickly achieves high accuracy on examples for which the class label, $\{ -1, +1 \}$, matches the corresponding majority group, $\{\textrm{inverted}, \textrm{not inverted}\}$, per the bias in the dataset.
By contrast, the accuracy on examples for which the class label does not match the majority group is very low in the initial stages of learning and increases much more gradually during training.
We surmise this happens because group membership (image inversion) is easier to recognize than class membership (CIFAR-100 binary label).
In the early stages of training, the model rapidly picks up on the easy-to-detect group membership signal as it provides the fastest way to reduce the model's loss.
In turn, this leads to bias amplification because the model makes predictions based on group membership signals whilst ignoring class membership signals.
As training progresses, the group membership signal loses value because it is not a perfect predictor of class membership (note that $\epsilon=0.3$).
Hence, the model starts to utilize more class membership signals as training progresses, which results in an increase in accuracy and a decrease in bias amplification.

\begin{figure}[t]
    \centering
    \includegraphics[width=0.51\textwidth]{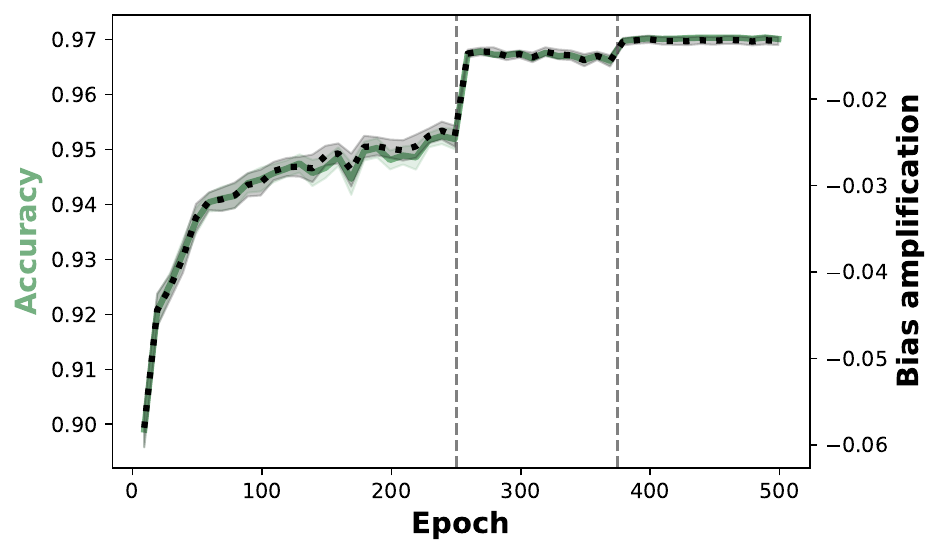}
    \hfill
    \includegraphics[width=0.46\textwidth]{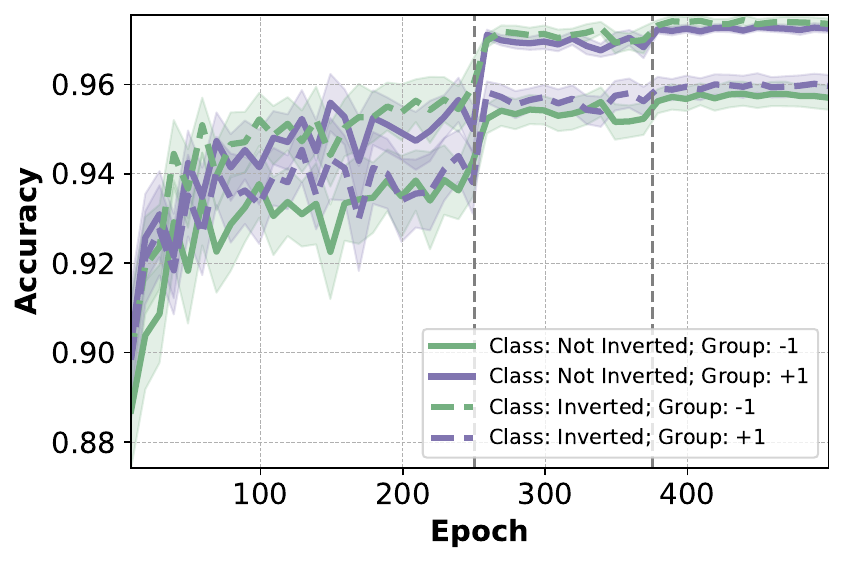}
    \caption{\textbf{Left:} Bias amplification and accuracy of ResNet-110 models during training on the CIFAR-100 dataset with a bias of $\epsilon=0.3$ in which \emph{the role of classes and groups is swapped} compared to the experiment in Figure~\ref{fig:biasamp_during_training}: the class label indicates whether or not an image is inverted, and the group label is determined based on the visual content of the image. \textbf{Right:} Accuracy of the models' per class-group combination. Shaded regions indicate the 95\% confidence intervals across 50 models. Vertical dashed lines indicate epochs at which the learning rate of the mini-batch SGD optimizer is decreased. }
    \label{fig:biasamp_during_training_swapped}
\end{figure}

We hypothesize that the early-stage bias amplification is due to group membership being easier to recognize than class membership in our setup.
To test this hypothesis, we perform an experiment in which we swap the role of the group and the class: \emph{i.e.}, the class label now represents whether or not the image is inverted and the group label depends on the object depicted in the CIFAR-100 image.
As before, we measure bias amplification during training and plot the results in the left pane of Figure~\ref{fig:biasamp_during_training_swapped}. 
The corresponding disaggregated accuracies are in the right pane of Figure~\ref{fig:biasamp_during_training_swapped}.
Indeed, we find that bias is dampened in the early stages of training as the model latches onto the easy-to-extract class membership signal first.
This bias dampening largely disappears in the later stages of training as the model starts to utilize group membership signals for recognition as well.

~\\
\paragraph{\bf RQ6: How does bias amplification vary as a function of the relative difficulty of recognizing class membership versus recognizing group membership?}~\\
Hitherto, we repeatedly observed that bias amplification may depend on the relative difficulty of recognizing class membership versus recognizing group membership:
as the group signal is easier to extract in our setup, models amplify bias more in early stages of training and/or when they have lower capacity.
This observation motivates a more detailed study of the effect of the relative difficulty of recognizing class membership and group membership on bias amplification.

\begin{figure}[t]
    \includegraphics[width=0.19\textwidth]{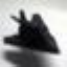}
    \includegraphics[width=0.19\textwidth]{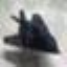}
    \includegraphics[width=0.19\textwidth]{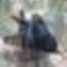}
    \includegraphics[width=0.19\textwidth]{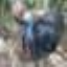}
    \includegraphics[width=0.19\textwidth]{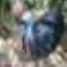}
\caption{Example of different amounts of overlay ($\eta = {0.0, 0.2, 0.5, 0.8, 1.0}$) for an example belonging to the ``airplane" class and ``bird" group. Only class information is visible when $\eta = 0.0$ (\textbf{left}); only group information is visible when $\eta = 1.0$ (\textbf{right}).}
\label{fig:overlay_example}
\end{figure}

To perform this study, we alter our problem setup in such a way that we can control the relative difficulty of class recognition and group recognition.
To do so, we abandon our image-inversion setup and, instead, create datasets that contain a convex combination of two CIFAR-10 images: a ``group image'' and a ``class image''.
By changing the weight of the convex combination, we can make the group image or the class image more prominent in the resulting image, thereby altering the difficulty of recognizing the class and the group.

We create the two groups, $a$ and $b$, by randomly choosing two CIFAR-10 classes that we sample group images from.
Similarly, we also randomly choose two CIFAR-10 classes to form the binary classification task (\emph{i.e.}, one class is the positive class and the other the negative class).
We use CIFAR-10 datasets because it has a larger class size than CIFAR-100.
Next, we create an example by sampling a class image, $\mathbf{I}_\textrm{class}$, from one of the two classes and a corresponding group image, $\mathbf{I}_\textrm{group}$, from one of the two group.
We linearly mix these two images:
\begin{equation}
\mathbf{I} = \eta \mathbf{I}_\textrm{group} + (1 - \eta) \mathbf{I}_\textrm{class},
\end{equation}
where $\eta \in [0, 1]$ is a mixing parameter and the final example $\mathbf{I}$ is assigned the label of $\mathbf{I}_\textrm{class}$.
Figure~\ref{fig:overlay_example} shows an example of the resulting examples  for different $\eta$ values.
As before, we assign positive examples to group $a$ with probability $0.5 + \epsilon$ or to group $b$ with probability $0.5 - \epsilon$. 
Negative examples are assigned group $b$ with probability $0.5 + \epsilon$, and to group $a$ with probability $0.5 - \epsilon$.

When $\eta=0$, this task reduces to classifying two classes from the standard CIFAR-10 images as the model cannot observe the group image at all: in this case, recognizing class membership is easy but it is not possible to identify group membership better than chance level.
Conversely, directly recognizing class membership is impossible when $\eta=1$ but recognizing group membership is easy in that setting.
Hence, $\eta$ provides a knob that facilitates varying the relative difficulty of recognizing group membership versus class membership.
We emphasize that $\eta$ can be varied separately from the bias parameter, $\epsilon$, which governs how likely it is that group membership predicts class membership correctly.

\begin{wrapfigure}{r}{0.5\textwidth}
  \includegraphics[width=0.5\textwidth]{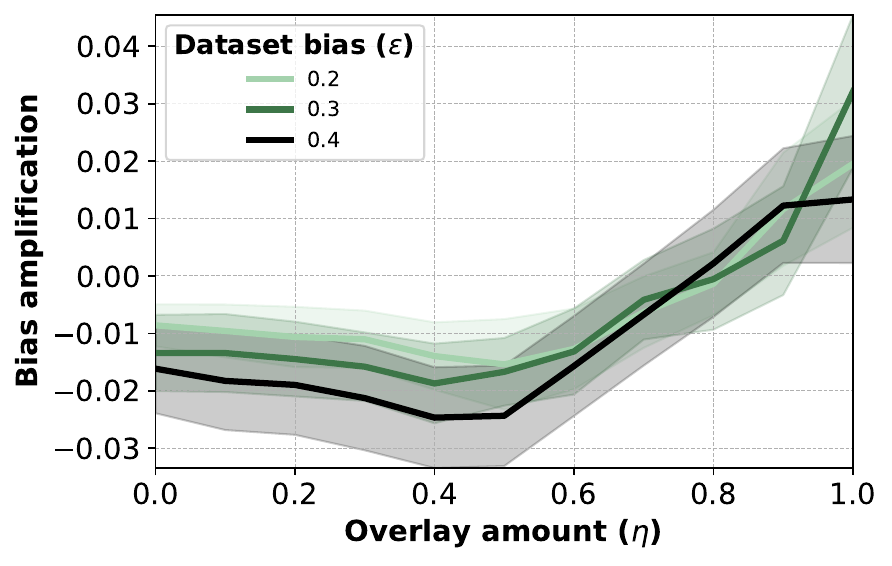}
  \caption{Bias amplification as a function of the relative difficulty of predicting class and group membership, $\eta$, for three different levels of bias, $\epsilon$. Recognizing class membership is easier for small $\eta$ values; recognizing group membership is easier for large $\eta$ values. Shaded regions indicate 95\% confidence intervals across 20 models.}
  \label{fig:overlay_biasamp_vs_relative_difficulty}
\end{wrapfigure}

Figure~\ref{fig:overlay_biasamp_vs_relative_difficulty} presents results of experiments in which we measure bias amplification as a function of the trade-off parameter, $\eta$, for different degrees of bias, $\epsilon$.
We follow the same setup as before and train ResNet-110 models in these experiments.
The results in the figure show that bias is dampened when it is relatively difficult to recognize group membership (\emph{i.e.}, when $\eta$ is low).
When $\eta$ increases past the point where group information is more visible than class information ($\eta=0.5$), however, the bias amplification starts to progressively increase and becomes positive for larger $\eta$. 
This observation provides additional evidence for the hypothesis that bias amplification depends heavily on the relative difficulty of recognizing group membership versus class membership.

\section{Related Work}
\label{sec:related_work}
This study is part of a larger body of work studying fairness and bias amplification in machine-learning models.

\paragraph{Fairness.}
Prior work has introduced a large number of formulations of fairness, each with their own tradeoffs in terms of group- and individual-level guarantees of equality or equity.
These formulations include equalized odds and equalized opportunity~\cite{hardt2016equality}, fairness through awareness~\cite{dwork2012fairness} or unawareness~\cite{grgic2016case,kusner2017counterfactual}, treatment equality~\cite{berk2021fairness}, and demographic parity~\cite{dwork2012fairness,kusner2017counterfactual}.
Measures associated with these fairness formulations include differences in accuracy~\cite{berk2021fairness}, differences in true or false positive rate~\cite{chouldechova2016fair,hardt2016equality}, and the average per-class accuracy across subgroups~\cite{boulamwini2018gendershades}.
These measures differ from bias amplification measures in that they focus on correlations in the model predictions, whereas bias amplification focuses on \emph{differences} between the correlations in the training data and those in the model predictions.
In other words, bias-amplification measures discern between bias that is adopted from the training data and bias that is amplified by the model; fairness measures make no such distinction.

\paragraph{Bias amplification.}
The study of bias amplification is of interest because it allows us to study how design choices in our models, training algorithms, \emph{etc.} contribute to bias in machine-learning models beyond biases in the training data~\cite{hooker2021moving}.
Prior work has measured bias amplification using generative adversarial networks~\cite{choi2020fair,jain2020imperfect}, by considering binary classifications without attributes~\cite{leino2019feature}, and by measuring correlations in model predictions~\cite{jia2020mitigating,zhao2017men}.
In our work, we use the $\biasamp$ measure from~\cite{wang2021directional}, which addressed shortcomings in prior work~\cite{zhao2017men}, to measure bias amplification.
Bias amplification has also been studied in the context of causal statistics~\cite{bhattacharya2007instrumental,middleton2016bias,pearl2010class,pearl2011invited,wooldridge2016instrumental}, but that line of work has remained  disparate from the study of bias amplification in machine learning.
Despite the plethora of prior work on measuring bias amplification, little is known on \emph{when and how} bias amplification arises in machine-learning models.
Our study is among the first to shed some light on the context under which bias amplification occurs. 

\paragraph{Calibration.}
Calibration of machine learning models is commonly considered desirable within traditional machine learning systems~\cite{Platt99probabilisticoutputs,naeini2015obtaining}. 
In recent years, calibration has been valued as a fairness guarantee as well: while calibration as a fairness measure is known to have weaknesses~\cite{corbett2018measure}, practitioners frequently find calibration-based approaches to be among the more useful fairness metrics in applications~\cite{bakalar2021fairness,obermeyer2019dissecting}.
This is especially true in systems that use model prediction scores (rather than hard labels), such as when building ranking and recommendation systems~\cite{steck2018calibrated}.
However, ~\cite{guo2017calibration} has shown that larger, newer neural networks are often less calibrated than their lower-capacity counterparts and tend to show overconfidence.
Our work is novel in identifying ways in which miscalibration and bias amplification are connected, suggesting that methods for mitigation may be shared between the two.

\section{Discussion}
\label{sec:discussion}
The results of our experiments shed light on the conditions under which bias amplification can arise in machine-learning models.
In particular, we find that bias amplification varies as a function of bias in the dataset, model capacity, training time, and the amount of training data. 
We also find that bias amplification depends on the relative difficulty of recognizing class membership and recognizing group membership.
This creates a predicament as the Bayes error of those two recognition tasks are generally beyond the control of the model developer.
Moreover, the model developer may not always be able to measure the difficulty of recognizing group membership empirically as doing so may involve developing a model that predicts sensitive attributes---something that model developers may want to avoid~\cite{keyes2018misgendering,larson2017gender,scheuerman2020how,wang2021directional}.

Although our study does not resolve this predicament, it may provide some useful best practices to mitigate bias amplification as much as possible during model development.
Specifically, our result suggests that there is value in using cross-validation to carefully select a model architecture, regularizer, and training recipe that minimizes bias amplification.
In other words, model developers may reduce bias amplification using the same tuning process that they routinely use to minimize classification error.\footnote{A potential downside is that such tuning does require access to sensitive attribute values, \emph{viz.} group-membership information.}
Our study provides intuitions for how some key levers available to the model developer, including model depth, model width, $\ell_2$ regularization, and learning rate schedule, can affect bias amplification.
Admittedly, our study does not provide a complete overview of how all relevant levers influence bias amplification; we intend to perform a more comprehensive investigation in future work.

\paragraph{Limitations.}
While our study provides useful insights and suggests best practices, it also suffers from several key limitations.
Importantly, our current study is limited to binary classification tasks in the image-recognition domain. 
Further work is needed to study bias amplification in the context of multi-class classification, regression, language modeling, and recommendation.
More work is also needed to understand if our observations generalize to settings in which there are more than two groups.
We note that in recommendation tasks, bias amplification may arise in more complex ways because such systems generally have a human-in-the-loop influencing the behavior of the system~\cite{bottou2013counterfactual}.

Another major limitation of our study is that it only studies bias \emph{amplification}.
While minimizing bias amplification is a valuable goal in itself, it may be insufficient for ensuring the machine learning predictions are unbiased and fair.
Optimizing models to reduce bias amplification may require tradeoffs between other fairness guarantees and other performance measures.
For example, achieving zero bias amplification in a system that is trained on data in which an advantaged group receives the preferred outcome at a higher rate than the disadvantaged group requires that the advantaged group continues to receive the preferred outcome at the higher rate, which implies demographic parity guarantees would not be satisfied. 
Reducing the bias of such a system (thus achieving negative bias amplification) may involve assigning qualified members of the advantaged group an unpreferred outcome, while giving similarly qualified members in the disadvantaged group a preferred outcome.
In turn, this would not satisfy group-level calibration guarantees.
Eliminating undesired biases altogether and ensuring fair, optimal system performance thus requires careful design of the entire pipeline from data collection to model deployment.

\bibliographystyle{ACM-Reference-Format}
\bibliography{references}

\end{document}